\documentclass{article}

\usepackage[preprint,nonatbib]{neurips_2022}
\usepackage{todonotes}
\usepackage[square,numbers]{natbib}
\usepackage{amsmath}
\usepackage{subfig}

\usepackage[utf8]{inputenc} %
\usepackage[T1]{fontenc}    %
\usepackage{hyperref}       %
\usepackage{url}            %
\usepackage{booktabs}       %
\usepackage{amsfonts}       %
\usepackage{nicefrac}       %
\usepackage{microtype}      %
\usepackage{xcolor}         %
\usepackage{siunitx}

\title{SolarDK: A high-resolution urban solar panel image classification and localization dataset}

\author{%
  Maxim Khomiakov$^{1,2}$ \\
   \And
   Julius Holbech Radzikowski$^{1*}$ \\
   \And
   Carl Anton Schmidt$^{1*}$ \\
   \And
   Mathias Bonde Sørensen$^{1*}$ \\
   \And
   Mads Andersen$^{1}$\thanks{Equal contribution} \\
   \And
   Michael Riis Andersen$^{1}$ \\
   \And
   Jes Frellsen$^{1}$ \\
   \And
   \vspace{0.5cm}
   $^{1}$\normalfont{Technical University of Denmark} \ \  $^{2}$\normalfont{Otovo AS}
  }

\begin{document}

\maketitle

\begin{abstract}
The body of research on classification of solar panel arrays from aerial imagery is increasing, yet there are still not many public benchmark datasets. This paper introduces two novel benchmark datasets for classifying and localizing solar panel arrays in Denmark: A human annotated dataset for classification and segmentation, as well as a classification dataset acquired using self-reported data from the Danish national building registry. We explore the performance of prior works on the new benchmark dataset, and present results after fine-tuning models using a similar approach as recent works. Furthermore, we train models of newer architectures and provide benchmark baselines to our datasets in several scenarios. We believe the release of these datasets may improve future research in both local and global geospatial domains for identifying and mapping of solar panel arrays from aerial imagery. The data is accessible at \href{https://osf.io/aj539/}{\texttt{https://osf.io/aj539/}}.
\end{abstract}

\section{Introduction}
The transition towards a more sustainable power generation has already begun. A core component of the energy mix in the future could very likely be solar panels also known as photovoltaics (PV) \cite{zappa2019100}. However, as the amount of deployed residential PV increases, the difficulty to balance the frequency of power for the Transmission System Operators (TSOs) becomes a harder task, as a greater part of the energy mix will be fluctuating due to meteorological variation caused by solar and wind. The ability to track solar adoption using remote sensing may benefit policy makers in their efforts to incentivize further green energy adoption, as identifying regions of lower solar adoption may be assisted using machine learning methods. Finally, there could be a viable motivation to understand the adoption of residential solar power, both in the sense of demographics studies as well as supporting greater adoption through hubs of installed PV---while some countries have good data and registration of residential installed PV, we often find it sparse, de-centralized and not easily accessible. Our contribution represents datasets spanning Denmark. We believe this dataset may contribute towards better models and generalization for the detection and localization of solar, and is as such a step in the right direction of the aforementioned goals. 

\section{Related works}
The early studies using machine learning for classification and localization of solar panels were commenced by \cite{bradbury2016distributed,malof2017deep,yu2018deepsolar}, who demonstrated a viable approach to identify and localize residential PV systems using an aerial or satellite imaging modality. While showing promise, subsequent studies found challenges upon deployment when performing inference across geographical domains \cite{huang2020deep}. The causes are considered to be a function of the image modality variability: ground sampling distance (GSD), angle of sampled image, time of day, or atmospheric based diffusion. This challenge begs the question of garnering labelled data from a variety of geographical locations in order to learn a better generalizable model. We identified three comparable datasets released in previous years: A segmentation dataset from the US \cite{bradbury2016distributed}, as well as two classification and segmentation datasets based in Germany \cite{mayer2020deepsolar} and France \cite{kasmi2022crowdsourced}. The US dataset spans four cities and consists of 19,863 polygons bounding the PV systems on residential houses. The German dataset consists of 107,809 images for classification and 4028 images for segmentation, while the more recent French based crowdsourced dataset \cite{kasmi2022crowdsourced}, contains 28,000 instances of PV systems with metadata including azimuth, surface, and slope, in addition around 20,000 annotated solar segmentations. 

\section{Dataset specification}

\begin{table}[tbp]
\caption{An overview over the presented datasets in this paper ($\dagger$ denotes that Gentofte Municipality is used for training and validation, and $\star$ denotes that Herlev Municipality is used as the test set). }
\label{tab:data_spec}
\centering
\begin{tabular}{@{}p{15mm}p{15mm}p{15mm}lll@{}}
\toprule
\textbf{Dataset}    & \textbf{Negatives} & \textbf{Positives} & \textbf{Area ($\mathrm{km}^2$)}   \\ \midrule
$\star$ Herlev  &    7,048            &       398       &        12.07                       \\
$\dagger$ Gentofte         &     15,489            &      482        &    25.70                            \\
BBR          &      -          &         104,397     &          3,853.02                     \\ \midrule
Total    &     22,537        & 105,344          &   3,890.79    & \\\bottomrule
\end{tabular}%
\end{table}

\begin{table}[tbp]
\caption{Number of examples per data split. The positive examples are identical for the classification and segmentation dataset.}
\centering
\label{tab:data_spec_train_test}
\begin{tabular}{llll}
\toprule
\textbf{Data split}    & \textbf{Negatives} & \textbf{Positives} & \textbf{Share of total (\%)}   \\ \midrule
Training  &    10,376            &       323       &        46                       \\
Validation         &     5,113            &      159        &    22                            \\
Test          &      7,048          &         398     &          32                     \\ \midrule
Total    &     22,537        & 880          &   100     \\\bottomrule
\end{tabular}%
\end{table}

In this paper, we present two types of datasets. Similarly to prior works of \citet{mayer2020deepsolar}, we provide a classification and a segmentation dataset. For the problem of classification, we provide manually labelled image instances from two urban municipalities in the Greater Copenhagen Region, while also providing an exogenous image dataset gathered from the Danish Building Registry (BBR), including images that span all over Denmark where a PV system is to be registered. The segmentation dataset covers the same instances manually labelled in the classification set. The images were manually labelled using Pigeon \cite{Pigeon}, while for the segmentation, we relied on AI-assisted annotation software by Toronto Annotation Suite \cite{torontoannotsuite}. We refer to Table \ref{tab:data_spec} for additional details and the Appendix for further examples and specifications of our labelling approach. SolarDK comprises images from GeoDanmark \cite{geodk} with a variable Ground Sample Distance (GSD) between $10\,\mathrm{cm}$ and $15\,\mathrm{cm}$, all sampled between March 1st and May 1st during 2021, containing 23,417 hand labelled images for classification and 880 segmentation masks, in addition to a set of about 100,000+ images for classification covering most variations of Danish urban and rural landscapes. We believe our datasets has the right combination of challenge as well as quality to further the development of a better generalizable method for the detection and localization of solar.

\section{SolarDK baselines}
We chose to focus our baselines revolving around the most widely used architectures found previously, with a particular emphasis on DeepSolarDE \cite{mayer2020deepsolar} and BBR to test geographical generalization. We introduce three scenarios for benchmarking SolarDK:

\begin{enumerate}
    \item Pre-trained models out of domain (i.e.\@ ImageNet).
    \item Pre-trained models out of domain with minority class augmentations from BBR.
    \item Pre-trained models of the same domain, yet different geographical region.
\end{enumerate}

The three blocks in Table \ref{tab:classification_benchmark_performance} are baselines corresponding to the three scenarios listed above. Except for the third block experiments, we trained all models five times with different random seeds. DeepSolarDK  follows the same approach as \citet{mayer2020deepsolar}, using the Inception V3 architecture \cite{szegedy2016rethinking}. The model tuning involved introducing augmentations such as horizontal and vertical flips, random perturbations to contrast and brightness, changes in initializations, as well as adjusting for the class imbalance by weighting the loss of the minority class higher, while also tuning hyperparameters like label smoothing, learning rate or dropout \cite{srivastava2014dropout}.

For the segmentation part, we produced two sets of baselines: One for pre-trained models out of the domain, and another by fine-tuning DeepSolarDE\cite{mayer2020deepsolar}. We compute all our classification and segmentation baselines using a threshold of 0.5, while using binary cross entropy loss to train the classification models, we apply the DICE-loss for the segmentation experiments.

The dataset was split into training/validation/test-splits where Gentofte municipality was used for training and validation, while Herlev municipality was used for testing purposes. The BBR dataset was used to augment the number of positive training samples in classification tasks, such that, instead of weighting the loss term of the minority class higher, we instead sample by random from the positive class set of BBR to gain an equal proportion of negatives and positive instances. These are the results illustrated by the asterisk in Table \ref{tab:classification_benchmark_performance}. We refer to the Appendix for further information on the datasets and the computed scenarios.

\subsection{Baseline metrics}
For the classification problem we use precision, recall and Cohen's $\kappa$, while for the segmentation problem we use mean Intersection over Union (mIoU) in addition to precision and recall.

\begin{table}[bp]
\centering
\caption{Classification baselines for the SolarDK testset. Each model was trained five times with different random seeds (except last block of models). * denotes runs augmenting the minority class using the BBR dataset, and bold-face denotes best mean performance(s). DeepSolarDE refers to a forward pass of \cite{mayer2020deepsolar} while DeepSolarDK is a fine-tuned version of \cite{mayer2020deepsolar} on the SolarDK dataset.}
\begin{tabular}{@{}llllll@{}}
\toprule
\textbf{Model}    & \textbf{Recall} & \textbf{Precision} & \textbf{Cohens ($\kappa$)}   \\ \toprule 
ConvNext &        0.60±0.04 &       \textbf{0.79±0.03} &        \textbf{0.66±0.02} \\
EfficientNet-b5 &       0.26±0.01 &       0.64±0.08 &        0.35±0.03 \\
EfficientNet-b7 &       0.35±0.05 &       0.71±0.02 &        0.45±0.04 \\
InceptionV3 &       0.34±0.18 &       0.56±0.38 &        0.55±0.04 \\
ResNet50 &       0.25±0.02 &       0.78±0.04 &        0.36±0.02 \\ 
ResNet101 &        0.58±0.40 &       0.49±0.39 &        0.41±0.21 \\
ResNet152 &       \textbf{0.65±0.16} &       0.51±0.28 &        0.49±0.14 \\\midrule
ConvNext* &       \textbf{0.65±0.07} &        0.70±0.06 &        \textbf{0.65±0.03} \\
EfficientNetb5* &        0.31±0.10 &        0.60±0.09 &        0.38±0.07 \\
EfficientNetb7* &       0.51±0.09 &       0.66±0.11 &        0.54±0.05 \\
InceptionV3* &       0.53±0.08 &       \textbf{0.73±0.09} &        0.58±0.05 \\
ResNet50* &       0.41±0.04 &       0.71±0.07 &        0.49±0.04 \\ 
ResNet101* &        0.41±0.10 &       0.65±0.03 &        0.46±0.08 \\
ResNet152* &       0.36±0.17 &       0.66±0.15 &          0.40±0.10 \\ \midrule
DeepSolarDE (inference)         &     0.42           &      0.17        &    0.21                            \\
DeepSolarDK &   \textbf{0.73}             &       \textbf{0.65}       &        \textbf{0.67}                       \\\bottomrule
\end{tabular}%
\label{tab:classification_benchmark_performance}
\end{table}

\begin{table}[tbp]
\caption{Segmentation baselines for the SolarDK testset. First group of runs were pre-trained using COCO train2017 and Pascal VOC, while the bottom section was pre-trained on German data \cite{mayer2020deepsolar}. The model with best mean performance(s) is marked with bold-face font.}
\centering
\begin{tabular}{@{}llllll@{}}
\toprule
\textbf{Model}    & \textbf{Recall} & \textbf{Precision} & \textbf{IoU}   \\\toprule
ResNet50-DeepLabV3Plus &       \textbf{0.81±0.03} &       0.86±0.01 &      \textbf{0.72±0.02} \\
ResNet101-DeepLabV3Plus &       0.79±0.05 &       0.86±0.02 &       0.70±0.03 \\
ResNet152-DeepLabV3Plus &       0.79±0.04 &       \textbf{0.88±0.03} &      0.71±0.02 \\
ResNet50-FPN &        0.80±0.03 &       0.87±0.03 &      \textbf{0.72±0.01} \\
ResNet101-FPN &       0.79±0.02 &       0.87±0.02 &      0.71±0.01 \\
ResNet152-FPN &      \textbf{0.81±0.06} &       0.87±0.05 &      \textbf{0.72±0.01} \\
ResNet50-PSPNet &       0.75±0.04 &       0.85±0.03 &      0.64±0.04 \\
ResNet101-PSPNet &       0.66±0.13 &      \textbf{0.88±0.05} &      0.61±0.07 \\
ResNet152-PSPNet &       0.72±0.05 &       0.85±0.04 &      0.63±0.02 \\
\midrule
DeepSolarDE (inference)         &     0.53           &      0.34        &    0.51\\
DeepSolarDK &   \textbf{0.85}             &       \textbf{0.75}       &        \textbf{0.62}                       \\\bottomrule
\end{tabular}%
\label{tab:segmentation_benchmark_performance}
\end{table}

\subsection{Results}
The results of our experiments are summarized in Table \ref{tab:classification_benchmark_performance} and \ref{tab:segmentation_benchmark_performance}. We see indications of better performance when using BBR as a substitute for minority class loss weighting. Similarly, we observe some of the best performance when using ConvNext \cite{liu2022convnet} as an encoder, while also seeing good performance from InceptionV3 \cite{szegedy2016rethinking}, in particular when using BBR augmentations. The model with best mean performance(s) is marked with bold-face font, but note that the intervals for several model are overlapping.   It is also worth to note the substantially subpar performance of DeepSolarDE \cite{mayer2020deepsolar} prior fine-tuning, which is an indication of imperfect geographical generalization.

\section{Discussion}
The baselines computed and gathered from our studies show that there is still a gap in solving the problem of identifying and localizing solar power. We do indeed verify there exists a known problem of deploying image based remote sensing neural networks on out of domain geographical regions \cite{wang2017poor,huang2020deep,davari2022generalizability}. We believe a way to solve this would be to call for the release of more annotated data, sampled across a variety of geographical regions, in addition to new developments in basic machine learning research that may allow for a better generalization throughout larger geographical regions. 

\section{Ethical considerations}
Privacy was our primary concern during our ethical considerations for publishing this dataset. Therefore we have chosen to reduce the personal identifiable information (PII) to a minimum, by not including the geospatial coordinates of the images or the PV systems. Additionally, we provide BBR metadata aggregated on a municipal level. 

\section{Conclusion}
In this paper we introduced new datasets for the detection and localization of solar panels, with a particular emphasis on solar panels deployed in urban environments. We have applied DeepSolarDE \cite{mayer2020deepsolar} for solar classification and localization tasks, in what may generally be perceived as a similar domain. We observe the failure for the best trained model from DeepSolarDE \cite{mayer2020deepsolar} to perform well on our datasets initially, however, upon using this model as a starting point for fine-tuning it very rapidly improves performance, on occasion in excess of models solely trained on the SolarDK dataset. This may indicate that DeepSolarDE \cite{mayer2020deepsolar} inherently does learn certain features that generalize as given by its out-of-the-box performance, while however not yet fully capable of being deployed across even fairly small geographical differences.  

\section{Acknowledgments}
We want to extend our thanks to Otovo AS for supporting this research.

\bibliographystyle{abbrvnat}
\bibliography{bibby}

\begin{thebibliography}{15}
\providecommand{\natexlab}[1]{#1}
\providecommand{\url}[1]{\texttt{#1}}
\expandafter\ifx\csname urlstyle\endcsname\relax
  \providecommand{\doi}[1]{doi: #1}\else
  \providecommand{\doi}{doi: \begingroup \urlstyle{rm}\Url}\fi

\bibitem[Bradbury et~al.(2016)Bradbury, Saboo, L~Johnson, Malof, Devarajan,
  Zhang, M~Collins, and G~Newell]{bradbury2016distributed}
K.~Bradbury, R.~Saboo, T.~L~Johnson, J.~M. Malof, A.~Devarajan, W.~Zhang,
  L.~M~Collins, and R.~G~Newell.
\newblock Distributed solar photovoltaic array location and extent dataset for
  remote sensing object identification.
\newblock \emph{Scientific data}, 3\penalty0 (1):\penalty0 1--9, 2016.

\bibitem[Davari~Majd et~al.(2022)Davari~Majd, Momeni, and
  Moallem]{davari2022generalizability}
R.~Davari~Majd, M.~Momeni, and P.~Moallem.
\newblock Generalizability in convolutional neural networks for various types
  of building scene recognition in high-resolution imagery.
\newblock \emph{Geocarto International}, 37\penalty0 (12):\penalty0 3565--3576,
  2022.

\bibitem[for Dataforsyning~og Infrastruktur(2022)]{geodk}
S.~for Dataforsyning~og Infrastruktur.
\newblock {GeoDanmark}.
\newblock
  \url{https://datafordeler.dk/dataoversigt/geodanmark-ortofoto/ortofoto-foraar-wms/},
  2022.

\bibitem[Germanidis(2020-2022)]{Pigeon}
A.~Germanidis.
\newblock {Pigeon}: Quickly annotate data on jupyter, 2020-2022.
\newblock URL \url{https://github.com/agermanidis/pigeon}.
\newblock Open source software available from
  https://github.com/agermanidis/pigeon.

\bibitem[Huang et~al.(2020)Huang, Bradbury, Collins, and Malof]{huang2020deep}
B.~Huang, K.~Bradbury, L.~M. Collins, and J.~M. Malof.
\newblock Do deep learning models generalize to overhead imagery from novel
  geographic domains? the xgd benchmark problem.
\newblock In \emph{IGARSS 2020-2020 IEEE International Geoscience and Remote
  Sensing Symposium}, pages 1476--1479. IEEE, 2020.

\bibitem[Kar et~al.(2021)Kar, Kim, Boben, Gao, Li, Ling, Wang, and
  Fidler]{torontoannotsuite}
A.~Kar, S.~W. Kim, M.~Boben, J.~Gao, T.~Li, H.~Ling, Z.~Wang, and S.~Fidler.
\newblock Toronto annotation suite.
\newblock \url{https://aidemos.cs.toronto.edu/toras}, 2021.

\bibitem[Kasmi et~al.(2022)Kasmi, Saint-Drenan, Trebosc, Jolivet, Leloux, Sarr,
  and Dubus]{kasmi2022crowdsourced}
G.~Kasmi, Y.-M. Saint-Drenan, D.~Trebosc, R.~Jolivet, J.~Leloux, B.~Sarr, and
  L.~Dubus.
\newblock A crowdsourced dataset of aerial images with annotated solar
  photovoltaic arrays and installation metadata.
\newblock \emph{arXiv preprint arXiv:2209.03726}, 2022.

\bibitem[Liu et~al.(2022)Liu, Mao, Wu, Feichtenhofer, Darrell, and
  Xie]{liu2022convnet}
Z.~Liu, H.~Mao, C.-Y. Wu, C.~Feichtenhofer, T.~Darrell, and S.~Xie.
\newblock A convnet for the 2020s.
\newblock In \emph{Proceedings of the IEEE/CVF Conference on Computer Vision
  and Pattern Recognition}, pages 11976--11986, 2022.

\bibitem[Malof et~al.(2017)Malof, Collins, and Bradbury]{malof2017deep}
J.~M. Malof, L.~M. Collins, and K.~Bradbury.
\newblock A deep convolutional neural network, with pre-training, for solar
  photovoltaic array detection in aerial imagery.
\newblock In \emph{2017 IEEE International Geoscience and Remote Sensing
  Symposium (IGARSS)}, pages 874--877. IEEE, 2017.

\bibitem[Mayer et~al.(2020)Mayer, Wang, Arlt, Neumann, and
  Rajagopal]{mayer2020deepsolar}
K.~Mayer, Z.~Wang, M.-L. Arlt, D.~Neumann, and R.~Rajagopal.
\newblock Deepsolar for germany: A deep learning framework for pv system
  mapping from aerial imagery.
\newblock In \emph{2020 International Conference on Smart Energy Systems and
  Technologies (SEST)}, pages 1--6. IEEE, 2020.

\bibitem[Srivastava et~al.(2014)Srivastava, Hinton, Krizhevsky, Sutskever, and
  Salakhutdinov]{srivastava2014dropout}
N.~Srivastava, G.~Hinton, A.~Krizhevsky, I.~Sutskever, and R.~Salakhutdinov.
\newblock Dropout: a simple way to prevent neural networks from overfitting.
\newblock \emph{The journal of machine learning research}, 15\penalty0
  (1):\penalty0 1929--1958, 2014.

\bibitem[Szegedy et~al.(2016)Szegedy, Vanhoucke, Ioffe, Shlens, and
  Wojna]{szegedy2016rethinking}
C.~Szegedy, V.~Vanhoucke, S.~Ioffe, J.~Shlens, and Z.~Wojna.
\newblock Rethinking the inception architecture for computer vision.
\newblock In \emph{Proceedings of the IEEE conference on computer vision and
  pattern recognition}, pages 2818--2826, 2016.

\bibitem[Wang et~al.(2017)Wang, Camilo, Collins, Bradbury, and
  Malof]{wang2017poor}
R.~Wang, J.~Camilo, L.~M. Collins, K.~Bradbury, and J.~M. Malof.
\newblock The poor generalization of deep convolutional networks to aerial
  imagery from new geographic locations: an empirical study with solar array
  detection.
\newblock In \emph{2017 IEEE Applied Imagery Pattern Recognition Workshop
  (AIPR)}, pages 1--8. IEEE, 2017.

\bibitem[Yu et~al.(2018)Yu, Wang, Majumdar, and Rajagopal]{yu2018deepsolar}
J.~Yu, Z.~Wang, A.~Majumdar, and R.~Rajagopal.
\newblock Deepsolar: A machine learning framework to efficiently construct a
  solar deployment database in the united states.
\newblock \emph{Joule}, 2\penalty0 (12):\penalty0 2605--2617, 2018.

\bibitem[Zappa et~al.(2019)Zappa, Junginger, and Van Den~Broek]{zappa2019100}
W.~Zappa, M.~Junginger, and M.~Van Den~Broek.
\newblock Is a 100\% renewable european power system feasible by 2050?
\newblock \emph{Applied energy}, 233:\penalty0 1027--1050, 2019.

\end{thebibliography}

\clearpage
\appendix

\section{Appendix}
\section{Experimental setup}
Our work encompasses the motivation to test three different aspects of the challenge at hand: The ability for the models to learn out of the box (w/o prior domain knowledge), the ability for in domain trained models to perform on a similar, but geographically different dataset, as well as assessing the value of using minority class augmentations from other geographical regions to boost performance. We used Adam as optimizer for all trained models.

\subsection{Models without prior domain knowledge}
All models shown in the first block of Table \ref{tab:classification_benchmark_performance} and Table \ref{tab:segmentation_benchmark_performance} have been trained on a subset of COCO train2017 and the 20 categories available in the Pascal VOC dataset. We apply minority class loss weighting of $\frac{|\text{Negatives}|}{|\text{Positives}|}$ to control for the class imbalance problem. The classification models with encoders ConvNext, ResNet and InceptionV3 were trained with a batchsize of 64 and a learning rate of 0.0001, while both of the EfficientNet encoders were trained with a batchsize of 32 and the same learning rate. 

\subsection{Models using BBR to augment minority class}
With all other settings being equal, we avoid doing any loss class imbalance weighting, and instead sample by random examples (outside of our Municipalities, Gentofte and Herlev) of the positive class from the BBR dataset such that the ratio between the negative and positive class is about 1:1. This is only performed for the training set. 

\subsection{Models with prior domain knowledge}
Prior training we compute a baseline simply doing a forward pass of the trained model through the SolarDK test set. Following such, we explore hyperparameters using the validation set and converge on the best set of models after a number of runs. The results from prior runs and the hyperparameters adjusted can be seen Table \ref{tab:appendix_valid_runs} below.

\subsubsection{DeepSolarDK}
In Table \ref{tab:appendix_valid_runs} we present the parameters used to identify the best performing model. From top left to right, the name of the configuration, optimizer chosen, batch size, label smoothing, learning rate, learning rate decay, recall, precision, Cohen's kappa and F1 score.

\begin{table}[!hp]
     \centering
     \caption{Table summarizing the prior runs for identifying hyperparameters for the DeepSolarDK model.}
     \begin{tabular}{c c c c c c c c c c}
        \toprule
         \multicolumn{10}{c}{\textbf{Validation performance}}\\ \toprule
         \textbf{Conf.} & \textbf{Opt.} & \textbf{BS}  & \textbf{LSR} & \textbf{LR} & \textbf{LRD} & \textbf{Recall} & \textbf{Pre} & \textbf{$\kappa$} & \textbf{F1} \\
         \hline
         \textsc{a} & Adam & 128 & 0.075 & 0.001 & 0.2 & \textbf{0.5882} & \textbf{0.6803} & \textbf{0.6195} & \textbf{0.6309} \\
         \textsc{b} & Adam & 32 & 0.075 & 0.0001 & 0.2 & 0.5294 & 0.6522 & 0.572 & 0.5844 \\
         \textsc{c} & Adam & 64 & 0 & 0.0001 & 0.3 & 0.4824 & 0.7257 & 0.5684 & 0.5795 \\
         \textsc{d} & Adam & 128 & 0.075 & 0.001 & 0 & 0.5412 & 0.5055 & 0.5063 & 0.5227 \\
         \textsc{e} & SGD & 32 & 0.075 & 0.0001 & 0.2 & 0.3882 & 0.3143 & 0.3232 & 0.3474 \\
         \textsc{f} & SGD & 64 & 0 & 0.0001 & 0.3 & 0.3294 & 0.2414 & 0.2507 & 0.2786 \\
          \textsc{g} & SGD & 128 & 0.075 & 0.0001 & 0.3 & 0.2706 & 0.2063 & 0.205 & 0.2341 \\
         \textsc{h} & SGD$^\circ$ & 32 & 0.075 & 0.0001 & 0.2 & 0.1176 & 0.1227 & 0.0915 & 0.1201 \\
         \midrule
         \textsc{i} & Adam$^\times$ & 64 & 0.075 & 0.0001 & 0.2 & 0.1984 & 0.1529 & 0.1488 & 0.1727 \\
         \textsc{j} & Adam$^\dag$ & 64 & 0.075 & 0.0001 & 0.2 & 0.21 & 0.1294 & 0.1395 & 0.1600 \\
         \textsc{k} & Adam$^\ddag$ & 64 & 0.075 & 0.0001 & 0.2 & 0.375 & 0.035 & 0.059 & 0.0645 \\
         \bottomrule
\end{tabular}
\label{tab:appendix_valid_runs}
\end{table}

\begin{table}[tbp]
     \centering
     \caption{Test performance comparison of the best three models based on their validation results (Table \ref{tab:appendix_valid_runs}) in comparison with DeepSolarDE prior fine-tuning on the SolarDK dataset.}
     \begin{tabular}{c c c c c}
        \toprule
         \multicolumn{5}{c}{\textbf{Test performance}}\\ \toprule
          \textbf{Conf.} & \textbf{Recall} & \textbf{Precision} & \textbf{Cohens} $\kappa$ & \textbf{F1} \\
         \midrule
         \textsc{b}  & 0.7337 & 0.6505 & 0.6717 & \textbf{0.6896} \\
         \textsc{c}  & 0.8139 & 0.5972 & 0.6729  & 0.6889 \\
         \textsc{a}  & 0.7232 & 0.6412 & 0.6613 & 0.6798 \\
         
         DeepSolarDE  & 0.4186 & 0.1667 & 0.2124 & 0.2384 \\
         \bottomrule
     \end{tabular}

     \label{tab:test_perf}
\end{table}
\section{Sample images from SolarDK}

\begin{figure}[!h]
    \centering
    \includegraphics[width=\linewidth]{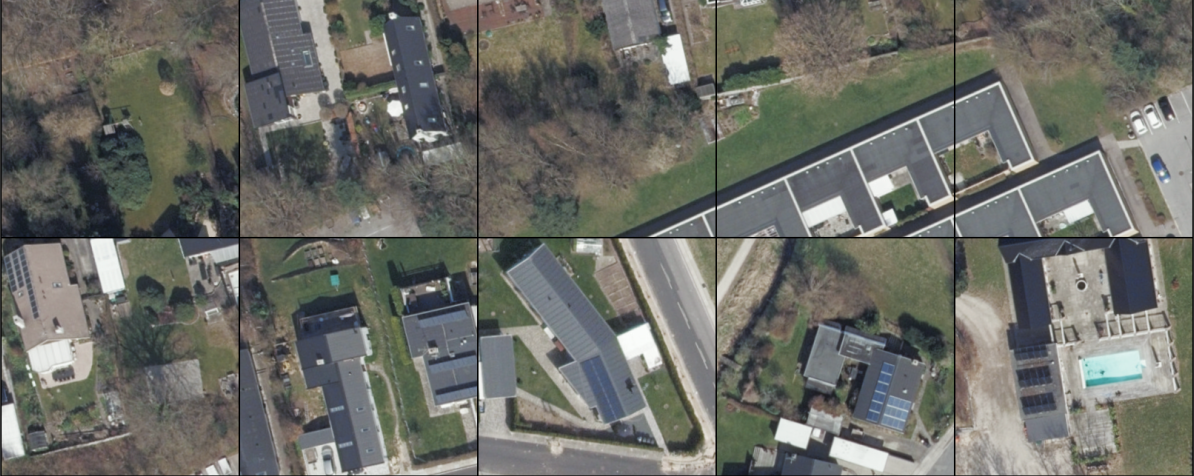}
    \caption{Illustrations of images from Herlev municipality (test set). Top row without PV, bottom row with PV.}
    \label{fig:herlev}
\end{figure}
\begin{figure}[ht]
    \centering
    \subfloat{{\includegraphics[width=6cm]{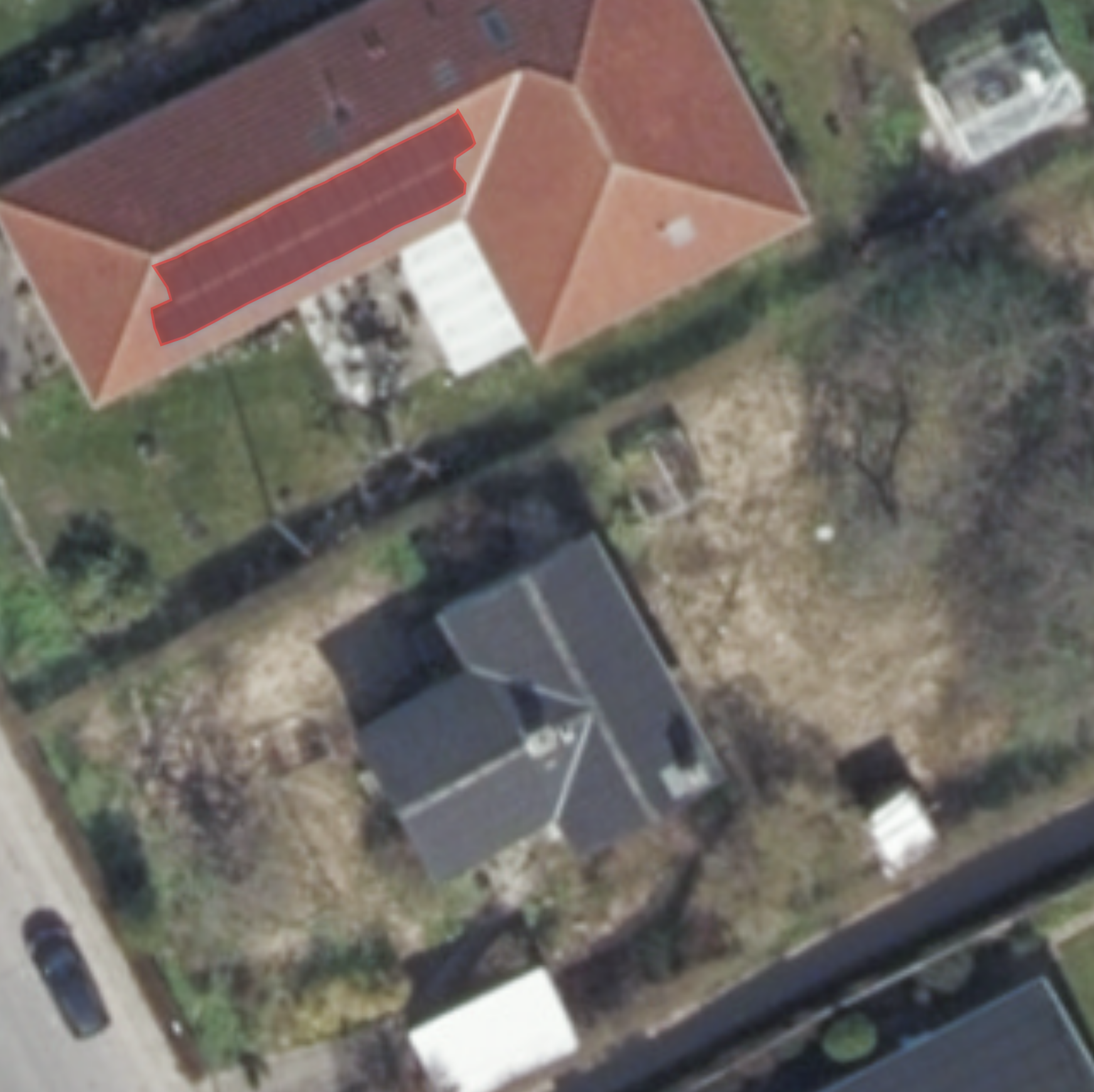} }}%
    \qquad
    \subfloat{{\includegraphics[width=6cm]{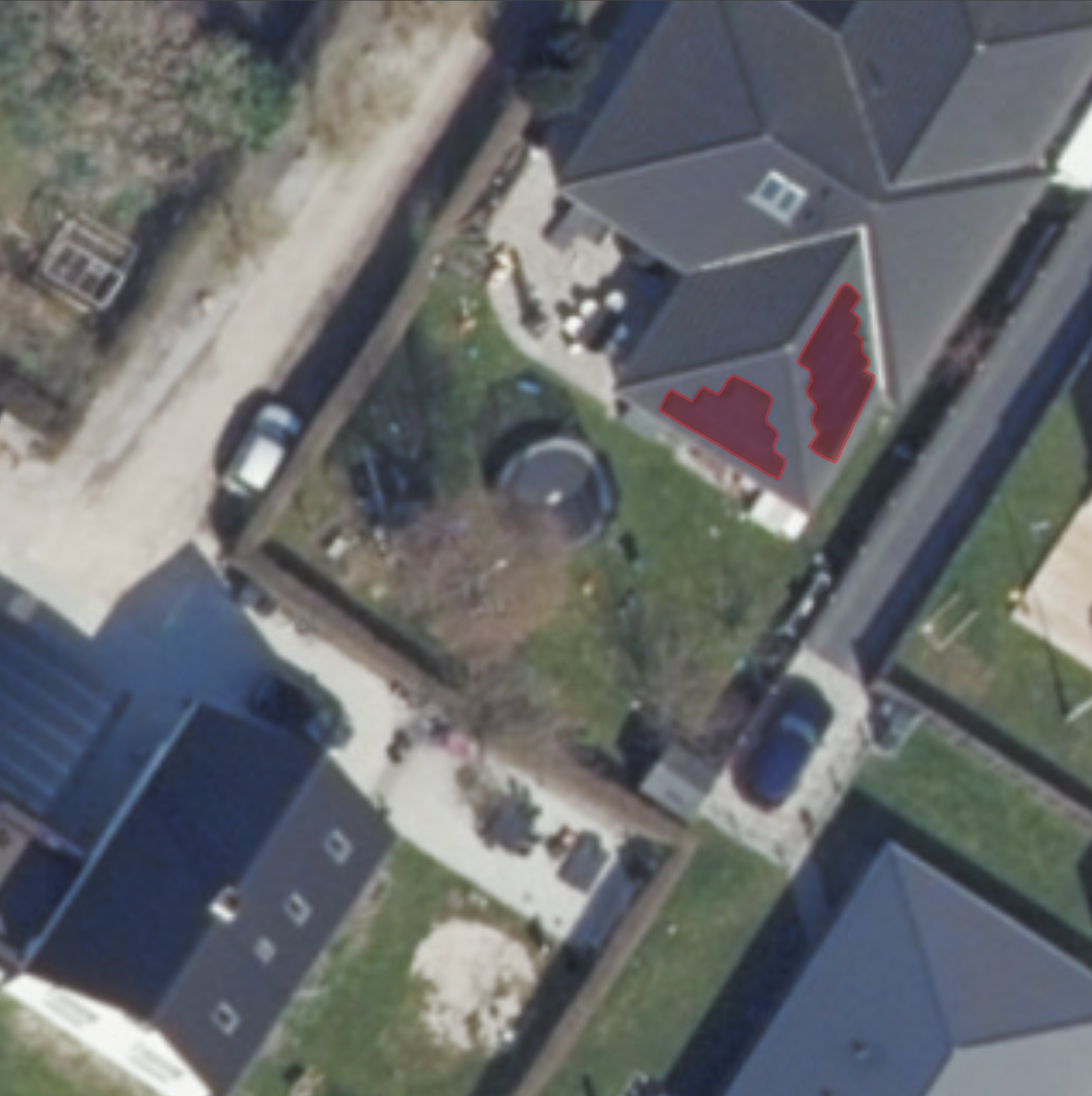} }}%
    \caption{Examples of segmentation masks (indicated by red).}
    \label{fig:segmentation}%
\end{figure}

\end{document}